\documentclass[a4paper, 10pt, conference]{ieeeconf}      % Use this line for a4
\usepackage{FG2018}

\FGfinalcopy % *** Uncomment this line for the final submission

\usepackage{graphics} % for pdf, bitmapped graphics files
\usepackage{graphicx}
\usepackage{subfigure}
\usepackage{amsmath}

\DeclareMathOperator*{\argmax}{arg\,max}

\def\FGPaperID{196} % *** Enter the FG 2018 Paper ID here

\title{\LARGE \bf
Attributes in Multiple Facial Images
}

\author{\parbox{16cm}{\centering
    {\large Xudong Liu  and Guodong Guo}\\
    {\normalsize
    Lane Department of Computer Science and Electrical Engineering , West Virginia University, Morgantown, WV 26506, USA\\
    xdliu@mix.wvu.edu, guodong.guo@mail.wvu.edu}}
}

\begin{document}

\IEEEoverridecommandlockouts\pubid{\makebox[\columnwidth]{978-1-5386-2335-0/18/\$31.00~\copyright{}2018 IEEE \hfill}
\hspace{\columnsep}\makebox[\columnwidth]{ }}

\ifFGfinal
\thispagestyle{empty}
\pagestyle{empty}
\else
\author{Anonymous FG 2018 submission\\ Paper ID \FGPaperID \\}
\pagestyle{plain}
\fi
\maketitle

%\ifFGfinal
%\thispagestyle{empty}
%\pagestyle{empty}
%\else
%\author{Anonymous FG 2018 submission\\ Paper ID \FGPaperID \\}
%\pagestyle{plain}
%\fi
%\maketitle

%%%%%%%%%%%%%%%%%%%%%%%%%%%%%%%%%%%%%%%%%%%%%%%%%%%%%%%%%%%%%%%%%%%%%%%%%%%%%%%%
\begin{abstract}

Facial attribute recognition is conventionally computed from a single image. In practice, each subject may have multiple face images. Taking the eye size as an example, it should not change, but it may have different estimation in multiple images, which would make a negative impact on face recognition. Thus, how to compute these attributes corresponding to each subject rather than each single image is a profound work. To address this question, we deploy deep training for facial attributes prediction, and we explore the inconsistency issue among the attributes computed from each single image. Then, we develop two approaches to address the inconsistency issue. Experimental results show that the proposed methods can handle facial attribute estimation on either multiple still images or video frames, and can correct the incorrectly annotated labels. The experiments are conducted on two large public databases with annotations of facial attributes.

\end{abstract}

%%%%%%%%%%%%%%%%%%%%%%%%%%%%%%%%%%%%%%%%%%%%%%%%%%%%%%%%%%%%%%%%%%%%%%%%%%%%%%%%
\section{INTRODUCTION}

Facial attributes are one of the most powerful descriptors for personality attribution \cite{penton2006personality}. In the area of computer vision, researchers have worked on the extraction and use of attributes in various tasks, such as object detection and classification \cite{szegedy2015going,simonyan2014very,he2016deep,zeng2016gated,szegedy2016rethinking}, as well as face recognition \cite{taigman2014deepface,sun2015deepid3,schroff2015facenet,zhou2015naive}. Facial attributes are beneficial for multiple applications, including face verification \cite{kumar2009attribute,song2014exploiting,berg2013poof} identification \cite{manyam2011two}, and face image search \cite{lei2011photo}. It is even shown that gender classification can be improved \cite{bekios2014robust} by exploiting the existence of dependencies among gender, age and other facial attributes.

Facial attributes are usually computed from a single face image, e.g., \cite{kumar2009attribute,liu2015deep,zhong2016leveraging,rudd2016moon}. However, we are interested in a related but different problem: How to compute the attributes given multiple face images of the same subject? In other words, our interest is to extract subject-based attributes, rather than the traditional single-image-based attributes.

In practice, it is quite common to capture multiple still images for each subject or to acquire a video of a subject with a number of image frames of the subject. Thus it is not rare to encounter the situation of having multiple still images or video frames of the same subject. Then it is quite natural to request a unique set of attributes about the subject given multiple face images, which is also beneficial for face recognition.

One possible way to derive the attributes from multiple images is to compute the attributes from each image and then get one common description of the subjects. This approach may raise an issue: Is there any inconsistency among the attributes computed from the single image? And if the inconsistency exists, how to address it?

\begin{figure}[h]
\begin{center}
\includegraphics[width=9cm]{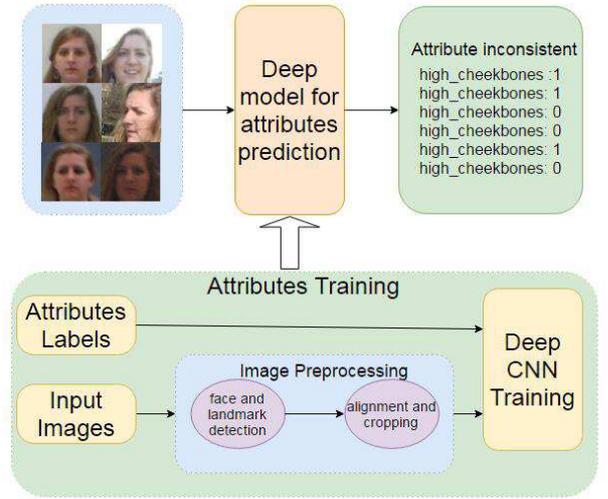}
\caption{\label{Fig.1}Overview of attributes inconsistency.}
\end{center}
\end{figure}

In this paper, we explore whether the inconsistency exists among the attributes computed from multiple face images of the same subject. The inconsistency can be caused by the variations in images, such as the face image quality changes. Then we develop methods to address the inconsistency.

Our main contributions include:

$\bullet$ We present a new problem, i.e., computing the subject-based attributes in contrast to the traditional single-image-based. The inconsistency problem is raised when multiple face images are given.

$\bullet$ Two approaches are developed to address the inconsistency issue among multiple images.

$\bullet$ Annotations of 40 attributes for two databases with a number of still images and video frames.

$\bullet$ Correct the incorrectly annotated attribute labels.

\section{RELATED WORK}

Kumar et al. \cite{kumar2009attribute}  employed face attributes for face verification using binary classifiers trained to recognize the presence or absence of describable visual appearance (face attributes). Due to the recent advances in GPUs and deep learning, Liu et al. \cite{liu2015deep} cascaded two CNNs, LNet for face localization and ANet for attributes prediction, which are fine-tuned jointly with attributes labels. They have achieved state-of-the-art performance for 40 face attributes prediction tested on CeleA and LFWA, respectively. Using \cite{liu2015deep}, Zhong et al. \cite{zhong2016leveraging} compared different features from different CNN layers and gained a better performance on face attributes prediction using the mid-level CNN feature. More recently, Rudd et al. \cite{rudd2016moon} proposed a novel mixed domain adaptive optimization network (MOON) for facial attribute recognition.
Almost all existing works focus on the estimation of face attributes from a single image. In contrast, a somewhat similar but different problem: how to compute the attributes given multiple images of the same subject? The multiple images can come from still images or video frames. When the attributes are computed from each single image, is there any inconsistency among them? If yes, how to address the inconsistency? All these questions will be addressed in the following.

%\addtolength{\textheight}{-3cm}   % This command serves to balance the column lengths

%%%%%%%%%%%%%%%%%%%%%%%%%%%%%%%%%%%%%%%%%%%%%%%%%%%%%%%%%%%%%%%%%%%%%%%%%%%%%%%%
\section{Inconsistence Measure}

We study the problem of face attribute inconsistency on multiple images from the same subject. Through  experiments, we found that there exits inconsistency. To quantify the inconsistency, we propose to measure the inconsistency degrees, named Inconsistence Measure (IM).

Suppose there are $L$ subjects, where $L=1,2,3,\ldots$. For the $l$-th subject, there are $N_{l}$ images, where $\sum\limits_{l=1}^{L}{N_{l}}=N$, $N=1,2,3,\ldots$. The $i$-th image of the $l$-th subject is denoted as $S_{l}^{i}$, where $\sum\limits_{i=1}{S_{l}^{i}}=N_{l}$. Here we define the binary classification:

\begin{equation}\label{classifer}
f_{j}(S^{i}_{l})=
\left\{
\begin{array}{l}
1,\quad if\ $j$^{th}\ attribute\ is\ true    \\
0,\quad otherwise,
%f\{j}(S^{i}_{l})=0$$
\end{array}
\right.
\end{equation}
where j denotes the attribute index, $j$=1,2,3,\ldots,40. Then the number of positive and negative prediction results can be caculated for each attribute from each subject.

\begin{equation}\label{n1}
  C_{l}^{j}(1)=\sum_{i=1}^{N_{l}}f_{j}(S^{i}_{l}).
\end{equation}

\begin{equation}\label{n0}
  C_{l}^{j}(0)=N_{l}-\sum_{i=1}^{N_{l}}f_{j}(S^{i}_{l}).
\end{equation}

Accordingly, a ratio to measure the portion between the positive and negative can be computed:

\begin{equation}\label{ratio}
R_{l}^{j}=\max \{C_{l}^{j}(1),C_{l}^{j}(0)\}/ N_{l},
\end{equation}
where $R \in $ [0.5,1]. If there are half positive and half negative attribute results, $R$ equals 0.5, which means that attribute has the most inconsistent issue, whereas that attribute is consistent when $R$ equals 1. $R$ is a basic measure for the inconsistency issue. To have a better measure, we re-scale and re-formula the ratio, as shown in (5) and (6).

\begin{equation}\label{ratio}
IM_{l}^{j}\prime = (R_{l}^{j}-0.5)/0.5*100.
\end{equation}

\begin{equation}\label{ratio}
IM_{l}^{j} = 100-IM_{l}^{j}\prime,
\end{equation}
where $IM \in $ [0,100]. The IM values can indicate the inconsistency degrees. The larger the IM, the more inconsistent the attribute. Accordingly, for the $j$-th attribute, IM can be calculated for all subjects:

\begin{equation}\label{ratio}
IM^{j} = \frac{1}{L}\sum_{l=1}^{L} IM_{l}^{j}.
\end{equation}

From equation (6), there will be no inconsistency when IM is zero. The higher IM indicates more inconsistency of an attribute. It is not difficult to understand that the attribute inconsistency will influence the face recognition performance for any attribute-based face recognition systems. For example, one person should have had the high cheekbone attribute, but it disappears because of occlusion reason during a short period of a video. Considering this problem, we propose two approaches to address the issue of attribute inconsistency in multiple images.

\section{ADDRESS THE INCONSISTENCY}

To address the inconsistency issue, we develop two different approaches. The first one is based on a probabilistic confidence, and the other is to consider the image quality. Both methods combine the estimation from multiple images, and eventually,  improve the attribute prediction performances at the subject level.

\subsection{Probabilistic Confidence Criterion}

Binary classifiers can be used for attribute recognition for each single image. Intuitively, an efficient classifier will not only be able to make the correct prediction but also has the highest confidence. Following this idea, we check the confidence of the result. The estimation of  facial attributes trained on the CelebA achieves a comparable performance to the state-of-the-art \cite{rudd2016moon}(see Section \uppercase\expandafter{\romannumeral5}), which means we have trained good deep features. Subsequently, binary classifier descriptors play an equally significant role in the final result. In this work, we deployed the random forest as the classifier.

We used 40 random forest models as the classifier descriptors. Random forest is made of plenty of decision trees. We generate each probability from these binary classifiers' outputs,
denoted as $P[f_{j}(S^{i}_{l})=1]$ and $P[f_{j}(S^{i}_{l})=0]$, and define confidence as:

\begin{equation}\label{ratio}
Confidence^{ij}_{l}= \left | P[f_{j}(S^{i}_{l})=1]-P[f_{j}(S^{i}_{l})=0] \right |,
\end{equation}
then, the representation of the $l$-th subject for the $j$-th attribute is computed as following:
\begin{equation}\label{ratio}
%\begin{displaymath}
Sub^{j}_{l}=\argmax_{i\in N_{l}} Confidence^{ij}_{l}.
%Sub^{j}_{l}=\argmax Quality^{ij}_{l}.
%\end{displaymath}
\end{equation}

As a consequence, we extract the most confident image representation for each subject. We then select the result from the highest confidence as the subject's attribute.

\subsection{Image Quality Criterion}

The face image quality may also cause the inconsistency issue in attribute recognition. We investigate 11 typical heuristic features for image quality assessment, which includes brightness \cite{haque2013real}, contrast, focus \cite{abaza2014design}, illumination, illumination symmetry, sharpness, compression \cite{wang2002no}, pose estimation \cite{zhu2012face}, eyes detection, mouth detection and face symmetry. We empirically assign weights to each individual measure and then add these scores to generate one final score for each image, where the weights are shown in TABLE I. Afterwards, we select the image with the highest scores for attribute recognition for each subject.
\begin{table}[htb]
\centering
\caption{\label{table3}The values for image quality weight.}
\begin{tabular}{cc|cc}
\hline
$\textbf{Feature}$   & $\textbf{Weight}$ & $\textbf{Feature}$   & $\textbf{Weight}$
%& $\textbf{brightness}$   &$\textbf{contrast}$  & $\textbf{focus}$ & $\textbf{illumination}$ & $\textbf{illumination symmetry}$ & $\textbf{sharpness}$ &$\textbf{compression}$ & $\textbf{pose}$ & $\textbf{eyes openness}$  & $\textbf{mouth closeness}$  & $\textbf{face symmetry}$
\\
\hline
\hline

$\textrm{brightness}$  & $\textrm{0.6}$ & $\textrm{compression}$ &0.7 \\
\hline
$\textrm{contrast}$  & $\textrm{0.6}$ & $\textrm{pose}$ &1.0 \\
\hline
$\textrm{focus}$  & $\textrm{0.8}$ & $\textrm{eyes openness}$ &0.5 \\
\hline
$\textrm{illumination}$  & $\textrm{1.0}$ & $\textrm{mouth closeness}$ &0.5 \\
\hline
$\textrm{illumination symmetry}$  & $\textrm{0.9}$ & $\textrm{face symmetry}$ &1.0 \\
\hline
$\textrm{sharpness}$  & $\textrm{0.8}$  \\
\hline
%&0.8 &1.0 &0.9 &0.8 &0.7 &1.0 &0.5 &0.4 &1.0   \\
%\hline
\end{tabular}
\end{table}
%\begin{equation}\label{ratio}
%Quality = 0.6*f_{1}+0.6*f_{2}+0.8*f_{3}+1.0*f_{4}+0.9*f_{5}+0.8*f_{6}+0.7*f_{7}+1.0*f_{8}+0.5*f_{9}+0.4*f_{10}+1.0*f_{11}$.
%\end{equation}

\subsection{Image Fusion}

Given the above approaches, through either the probabilistic confidence or image quality criterion, we can improve performance by combining more representations. Taking probabilistic confidence as an example, we select the image that has the highest confidence. Furthermore, we select and combine the top 3 or 5 confidences for each subject. We use the majority voting as the final prediction. Eventually, the attribute recognition performance can be improved by such a fusion. The same strategy can be applied to the image quality based as well.

\section{Experiment}

\begin{figure*}[htb]
\begin{center}
\subfigure[]{\includegraphics[width=0.8\textwidth,height=0.25\textheight]{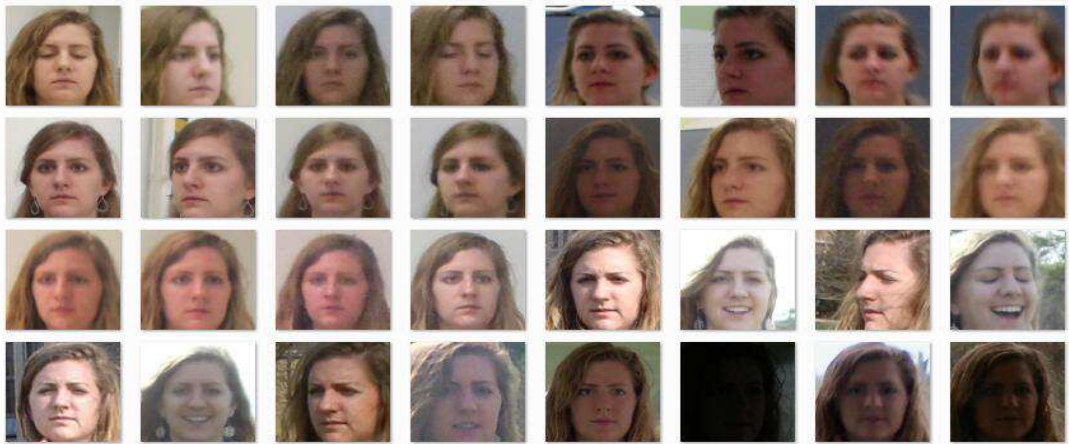}}
\subfigure[]{\includegraphics[width=0.6\textwidth,height=0.4\textheight]{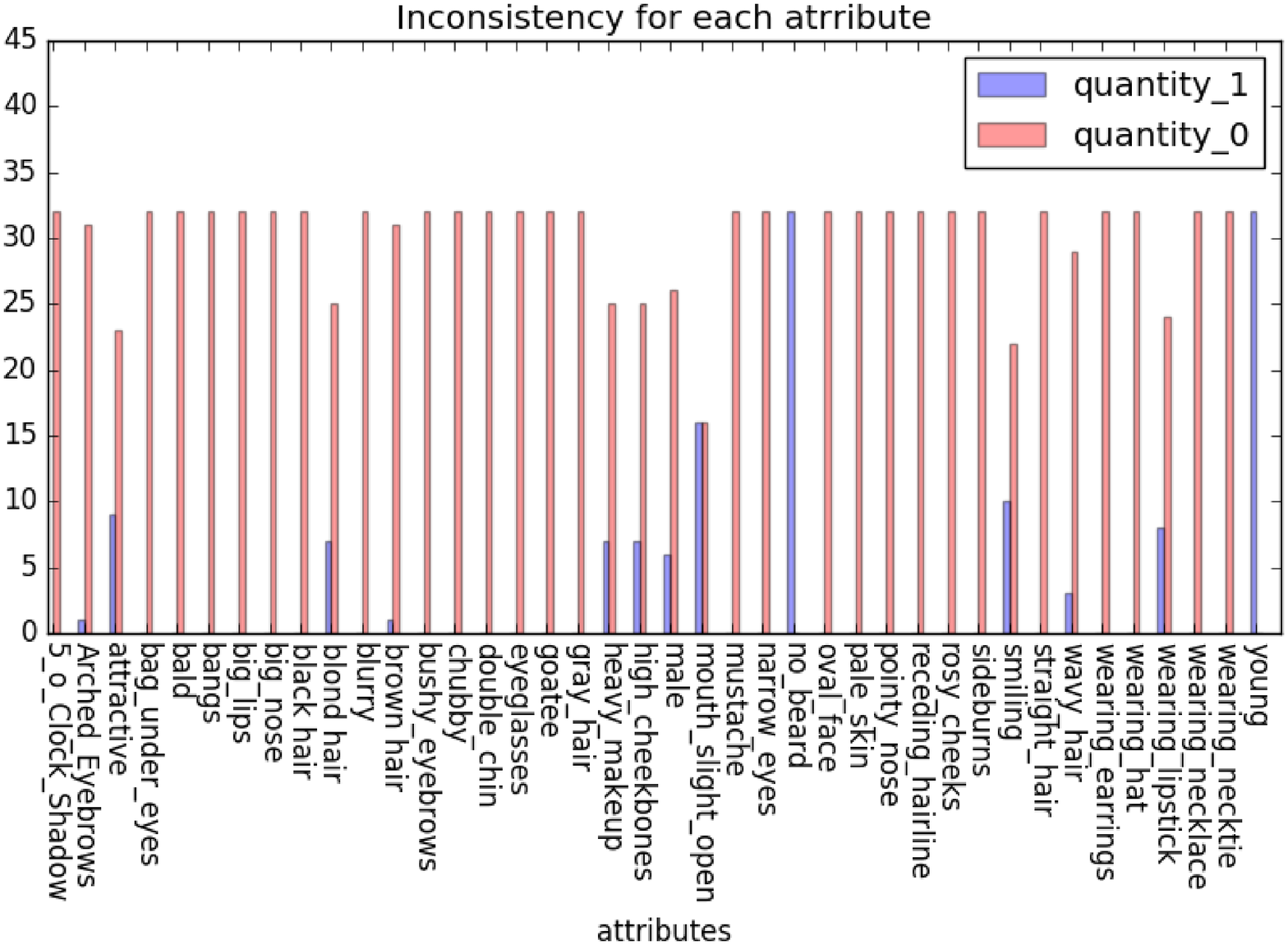}}
\subfigure[]{\includegraphics[width=0.3\textwidth,height=0.4\textheight]{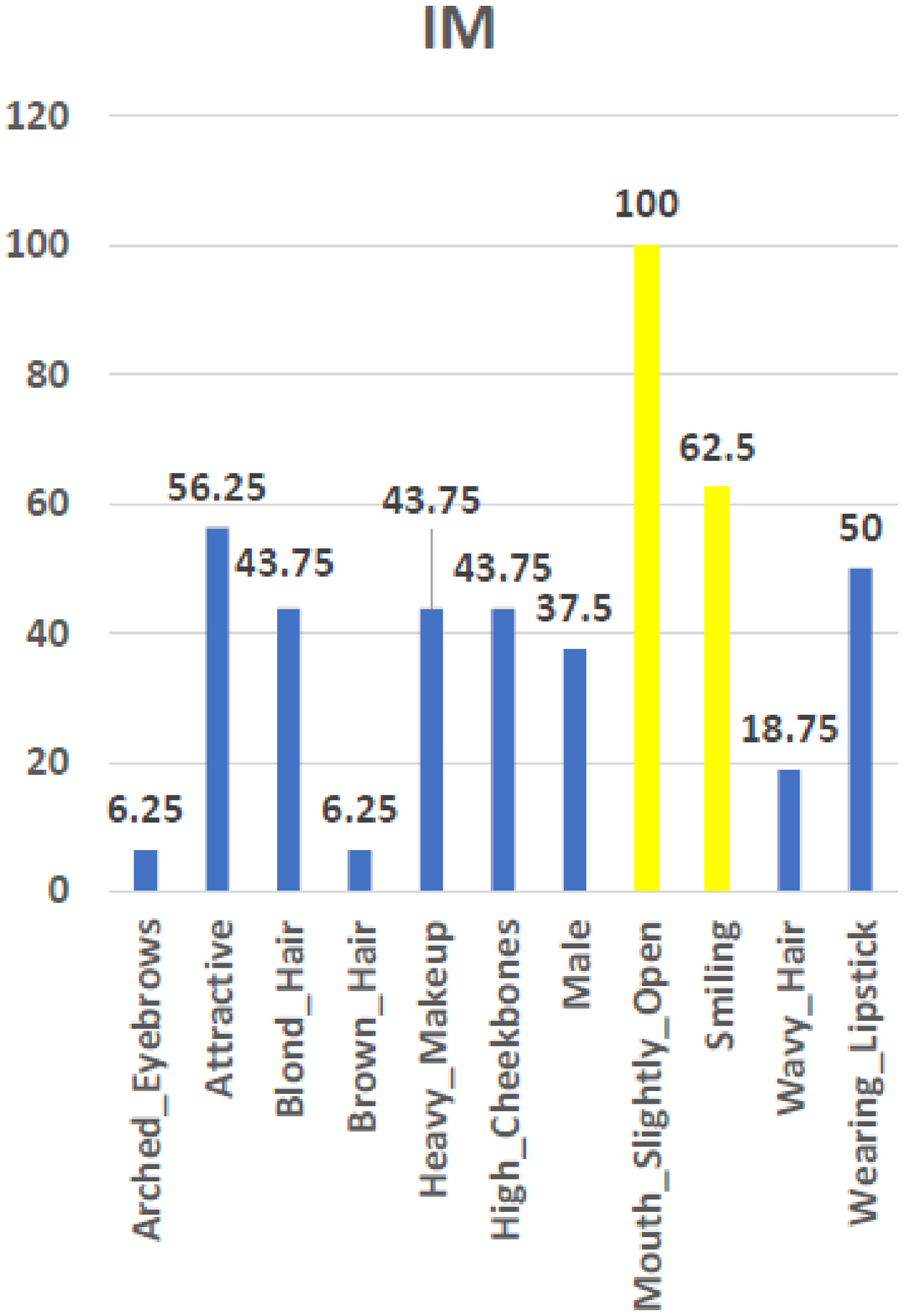}}
\caption{\label{Fig.2}(a) Examples of one identity from PaSC still images and corresponding to the 40 attributes recognition results (b) as well as the generated IM (c). The images show many variations in (a), facial attributes excluding attractive, mouth slight open and smiling (yellow bar in (c)) which are depended on each image. However, attributes, like Blond hair, High cheekbones, Male, which should be consistent but experimental results show inconsistent. (c) is the IM results, the higher IM the inconsistency is more serious, e.g. IM for Month\_Slightly\_Open is 100 which means numbers of 0 and 1 are equal.(best viewed in color)}
\end{center}
\end{figure*}

\subsection{Data}

In this work, we employ CelebA database for face attributes training. There are 200,000 images in CelebA, including 10,000 identities, each of which contains around 20 images. For each image, 40 face attributes are labeled, in other words, 8,000,000 attributes are provided in total on this database.

In order to measure the subject level facial attributes, we annotated 40 attributes on two datasets. One has 293 identities from PaSC \cite{beveridge2013challenge}. There are 293 identities from PaSC testing dataset, including 9376 still images (about 32 images per subject), and 2802 videos (approximately ten videos per person and 100 frames per video). Another dataset is COX \cite{huang2015benchmark}, which has 1,000 subjects, 3 videos  captured for each subject with 3 different camcorders. An interactive tool for annotating facial attributes was developed, which displayed multiple face images from the same subject. A rater was asked to check each attribute. Each subject was labeled by 3 volunteers. In order to obtain the subject level labels, we finalized the labels using a majority voting to get the unique result for each attribute. Therefore, 1293 subjects with 51720 facial attributes are used in our experiments.

\subsection{Deep Training for Facial Attributes Recognition}

Liu et al. \cite{liu2015deep}, released the labeled CelebA to the public and they reached 87\% accuracy over 40 attributes using LNets+ANet. Zhong et al. \cite{zhong2016leveraging} proposed to leverage the mid-level representations from off-the-shelf architecture to tackle the attribute prediction problem for fasces in the wild. They deployed different deep architecture, but both of them construct SVM as attributes classifier. Authors in \cite{rudd2016moon} proposed a novel mixed objective optimization network (MOON) to handle the imbalanced data and advanced state of the art in facial attributes recognition.

We deploy the GoogLeNet \cite{szegedy2015going} architecture for training deep model and random forest for the classifier. Sigmoid cross-entropy is used as the loss function, the learning rate is $10^{-4}$ with a polynomial decay. Features are extracted from FC layer,  and then we trained 40 random forest classifiers for attribute estimation using the deep features. Following the protocol \cite{liu2015deep}, which has three separated  parts: 160,000 images of 8,000 identities are used for deep training, and the images of another 20,000 of 1,000 identities are employed to train the random forest. The remaining 20,000 images of 1,000 identities are used for testing.

In addition, random forest not only can mostly avoid over-fitting compared to the single decision tree but also does not need tons of parameters to tune as SVM. For these reasons, we deploy random forest algorithm as our classifier to estimate the attributes. Random forest is much faster than SVM in our practice. After optimization of these models, we have achieved 87.7\% accuracy over the 40 facial attributes, which is comparable to the state-of-the-art.

\subsection{Mutiple Still Images and Videos on PaSC}

Even though both still images and videos \cite{beveridge2013challenge} are from several locations (inside buildings and outdoors), pose angles, different distances, as well as numbers of sensors, some kind of intrinsic attributes, e.g., gender, nose size, hair color, face shape, narrow eyes, pale skin, should be consistent at least for years. In addition, many attributes, such as, arched eyebrows, bald, bangs, chubby, double chins, goatee, high cheekbones, mustache, receding hairline, sideburns, hair shape, wearing earrings, wearing necklace, wearing necktie, also should not change for each person during a short time period. However, it would be challenging for face recognition when these facial attributes become inconsistent.

For still images, i.e., several images for the same subject. We compute the IM for each subject with each of the 40 attributes, using (6). One subject example for IM is shown in Fig. 2. We then concatenate the holistic still images using (7) and the whole IM are given in TABLE II.

\begin{table}[htb]
\centering
\caption{\label{table1}Inconsistence Measure on PaSC.}
\begin{tabular}{c|cc}
\hline
$\textbf{Attributes}$  & $\textbf{Still images}$   &$\textbf{Video frames}$  \\
\hline
\hline
$\textrm{Arched\_Eyebrows}$  & $\textrm{28.81}$     &31.30     \\
\hline
$\textrm{Attractive}$  & $\textrm{5.67}$     &5.51     \\
\hline
$\textrm{Bangs}$  & $\textrm{12.71}$     &22.89     \\
\hline
$\textrm{Big\_Nose}$  & $\textrm{0.53}$     &0.34     \\
\hline
$\textrm{Bushy\_Eyebrows}$  & $\textrm{0.28}$     &0.31     \\
\hline
$\textrm{Eyeglasses}$  & $\textrm{63.71}$     &60.14     \\
\hline
$\textrm{Heavy\_Makeup}$  & $\textrm{1.98}$     &1.36     \\
\hline
$\textrm{High\_Cheekbones}$  & $\textrm{47.83}$     &50.12     \\
\hline
$\textrm{Male}$  & $\textrm{0.19}$     &0.38     \\
\hline
$\textrm{Pointy\_Nose}$  & $\textrm{0.21}$     &0.53     \\
\hline
$\textrm{Straight\_Hair}$  & $\textrm{0.17}$     &0.13     \\
\hline
$\textrm{Wearing\_Lipstick}$  & $\textrm{63.52}$     &42.50     \\
\hline
$\textrm{Young}$  & $\textrm{0.23}$     &0.19     \\
\hline
\end{tabular}
\end{table}

After IM generated, the inconsistency issue is clear in TABLE II. We addressed the inconsistency as given in Section \uppercase\expandafter{\romannumeral4}.
As a consequence, we obtain a unique result for each attribute, and achieve 85.6\% and 83.0\% over 40 attributes based on the two criteria, respectively, as shown in TABLE \uppercase\expandafter{\romannumeral3}.

We can also apply the strategies to video frames. The difference is that while each video is considered as a subject for the video experiments, rather, each identity is denoted as one subject for still image experiments. There are several videos from the same identity in PaSC; it makes no sense if we simply combine different videos even they are from the same identity, because different videos should have their inconsistency issues. As the preceding analysis, we compute the highest confidence and the highest image quality, respectively. Afterward, we can provide unique results over 40 attributes for each video. Ultimately, the performance of videos reached 84.8\% and 83.8\% based on probabilistic confidence and image quality assessment, respectively, as shown in TABLE \uppercase\expandafter{\romannumeral3}.

\begin{table}[htb]
\centering
\caption{\label{table2}PERFORMANCE AFTER SELECTION.}
\begin{tabular}{c|cc}
\hline
$\textbf{}$  & $\textbf{Confidence}$   &$\textbf{Image Quality}$  \\
\hline
\hline
$\textrm{PaSC Still}$  & $\textrm{85.6\%}$     &83.0\%     \\
\hline
$\textrm{PaSC Video}$  & $\textrm{84.8\%}$     &83.8\%    \\
\hline
\end{tabular}
\end{table}

\subsection{Videos on COX}

The COX \cite{huang2015benchmark} consists of 1,000 subjects and three videos for each subject. We focus on the videos, which contain several frames, and demonstrate the attribute inconsistency issue.

\par We first compute the inconsistency from the entire video database on COX, and the IM is calculated as shown in TABLE \uppercase\expandafter{\romannumeral4}. Except for some attributes that exist for a short time, such as Mouth Slightly Open, Smiling, we are still able to find seven facial attributes that are inconsistent. As a result, we deploy our approaches to define these attributes on each video.

\begin{table}[htb]
\centering
\caption{\label{table3}Inconsistence Measure (IM) on COX.}
\begin{tabular}{c|ccc}
\hline
$\textbf{Attributes}$  & $\textbf{Cam1}$   &$\textbf{Cam2}$  & $\textbf{Cam3}$ \\
\hline
\hline
$\textrm{Attractive}$  & $\textrm{13.15}$     &9.40  &6.97   \\
\hline
$\textrm{Bangs}$  & $\textrm{3.6}$     &0 &0.68     \\
\hline
$\textrm{Eyeglasses}$  & $\textrm{0.34}$     &0.02 &0     \\
\hline
$\textrm{High\_Cheekbones}$  & $\textrm{17.68}$     &18.72 &24.25     \\
\hline
$\textrm{Male}$  & $\textrm{0.51}$     &0.32 &0.32    \\
\hline
$\textrm{Wearing\_Lipstick}$  & $\textrm{1.71}$     &0.13 &1.29     \\
\hline
$\textrm{Young}$  & $\textrm{0.32}$     &0.56 &0.02     \\
\hline
\end{tabular}
\end{table}

Similar to PaSC videos, we use the binary decision confidence for each frame in each video, before the final decision. For each video, we search the most confident frame for the attribute estimation.
On the other hand, there are some variations in each video clip, such as illumination, pose variation, blur, etc. As a consequence, we adopt the measured approach for image quality as we described in Section \uppercase\expandafter{\romannumeral4}. After the quality ranking, the highest quality image frame in each video is taken as input for attribute prediction. The accuracies over 40 attributes from all three camcorders videos are shown in Fig.3.

\begin{figure}
  \centering
  \includegraphics[width=9cm]{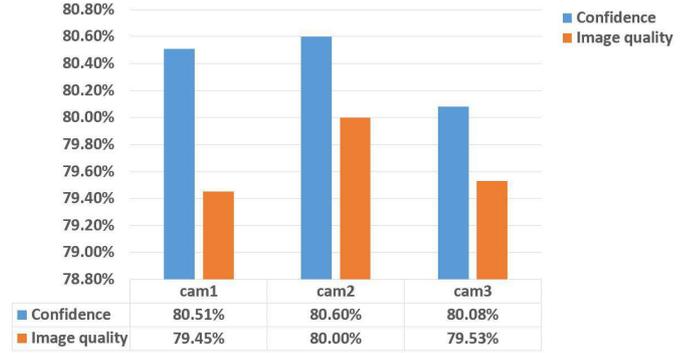}
  \caption{\label{fig.3} Attributes accuracy on COX}
\end{figure}

\subsection{Results from Fusion}

As discussed in Section \uppercase\expandafter{\romannumeral4}, we not only consider the best representation for each subject, but also improve the performance with fusion. From the probabilistic confidence on PaSC, we find out the best performance (86.0\%) comes up when we consider the top 3 for fusion. Additionally, given image quality, we gain the best performance with the fusion of top 5, as shown in Fig.4.

\begin{figure}[htb]
  \centering
  \includegraphics[width=9cm]{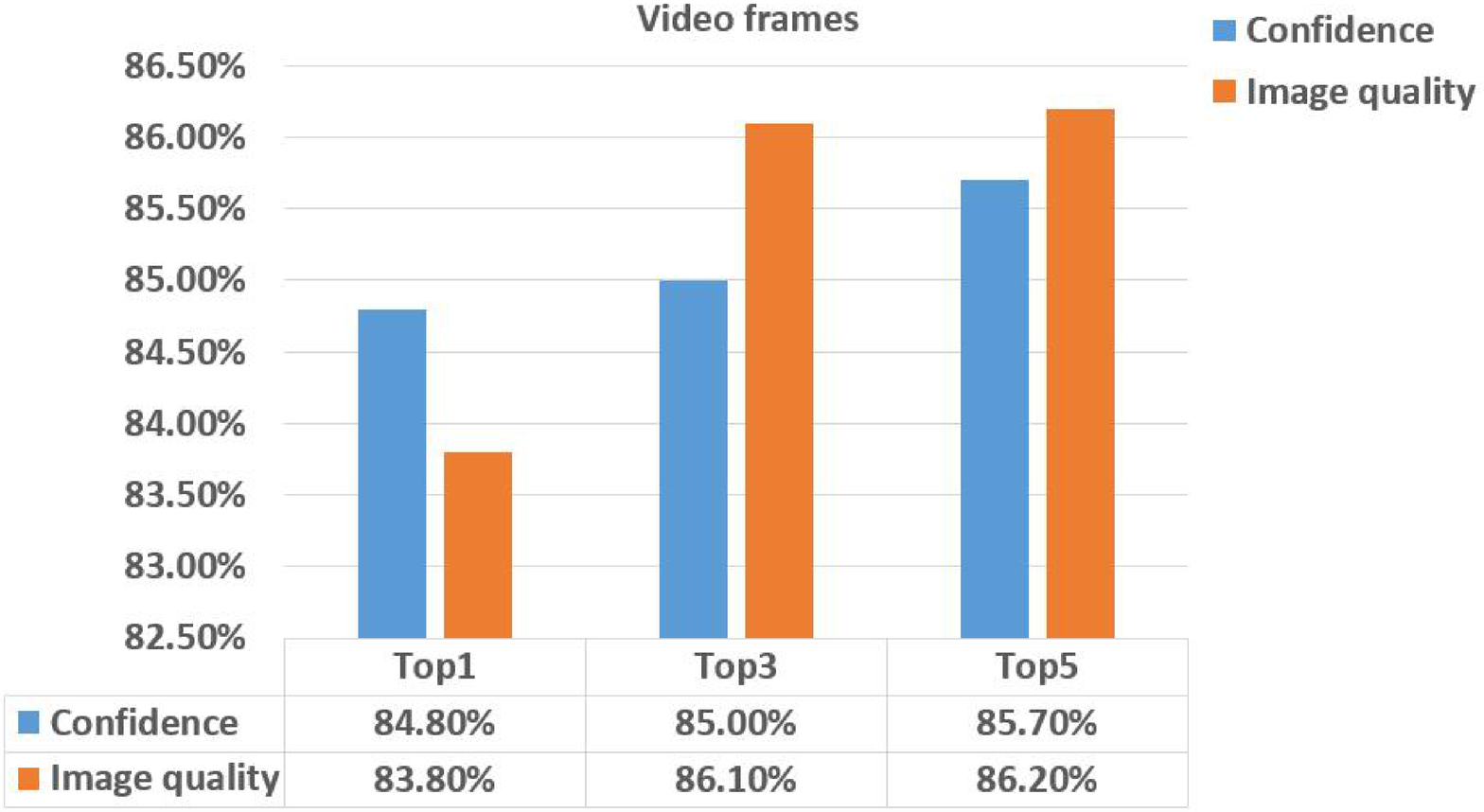}
  \includegraphics[width=9cm]{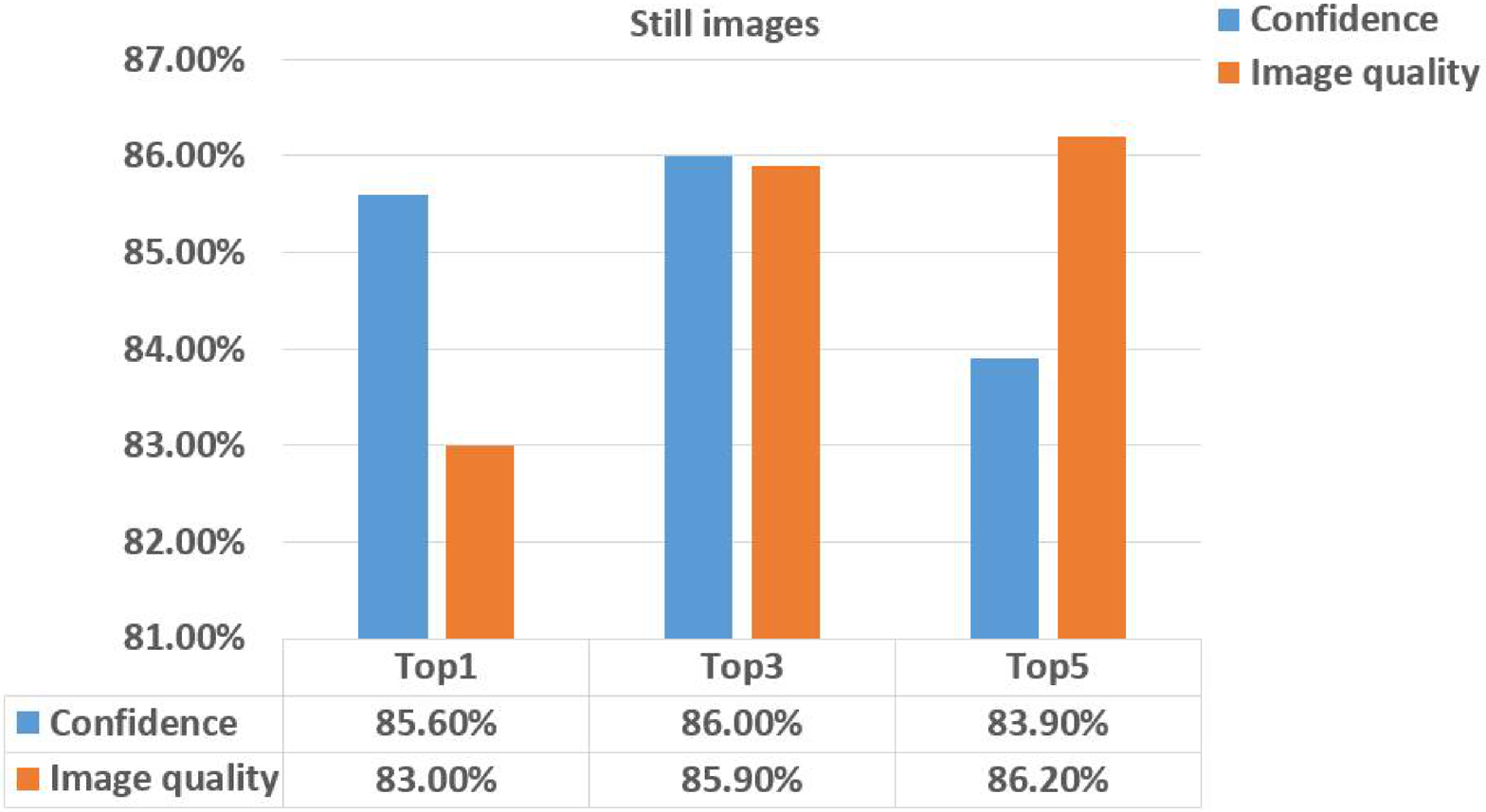}
  \caption{\label{fig.4} Fusion images from confidence and image quality perspective. Better performance both stills and videos on PaSC compared to top1.}
\end{figure}

From the experiments, we found that it is not true to get a better result with more images to combine for probabilistic confidence criteria. When considering more images, the chance that images with weak confidence dominate the result is increasing. While the top 3 can be fused to achieve the best performance based on probabilistic confidence. For quality assessment, we can see in Fig.5, the images keep a high quality through top 1 to 5. Therefore, the more images we are taking, the better performance we achieve. After fusion experiments,  the accuracy is improved to 86.2\% both Still and videos on PaSC.

%Form the experiments, we found that it is not true to get a better result with more images to combine. When considering more images, the probabilistic confidence value suffers quickly with more images though experiment results, which means the chance that images with weak confidence dominate the effect is increasing. Consequently, there is no doubt that the top 3 can achieve the best performance based on probabilistic measure. In contrast, we can see Fig.5. The images still keep a high quality through top 1 to top 5. Therefore, the more images we are taking, the better performance we achieve. After experiments and analysis, we have improved and obtained 86.2\% both Still and video on PaSC dataset.

\begin{figure}
  \centering
  \includegraphics[width=9cm]{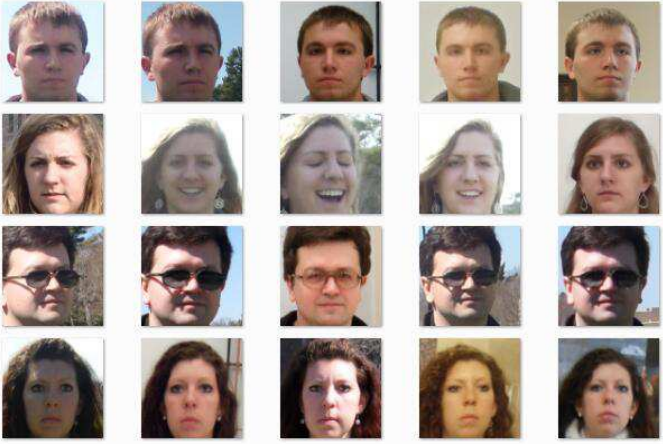}
  \caption{\label{fig.5} Image quality ranking examples on PaSC. top 1 to top 5 from left to right.}
\end{figure}

\subsection{Correct the Incorrectly Annotated Labels on CelebA}

There are 1,000 identities on CelebA testing set. We explore whether there are also inconsistencies for attributes. Similar procedures as we described on PaSC and COX datasets, we first extract the deep feature and proceed the attributes prediction based on each identity. Computed by (7), the IM values are generated as shown in TABLE \uppercase\expandafter{\romannumeral5}.

\begin{table}[htb]
\centering
\caption{\label{table4}Inconsistence Measure on CelebA.}
\begin{tabular}{cc|cc}
\hline
$\textbf{Attributes}$  & $\textbf{IM}$   &$\textbf{Attributes}$  &$\textbf{IM}$  \\
\hline
\hline
$\textrm{Attractive}$  & $\textrm{7.06}$    &$\textrm{High\_Cheekbones}$ &24.39     \\
\hline
$\textrm{Bangs}$  & $\textrm{2.39}$     &$\textrm{Male}$ &6.95  \\
\hline
$\textrm{Big\_Nose}$  & $\textrm{0.07}$    &$\textrm{Mouth\_Slightly\_Open}$ &65.47     \\
\hline
$\textrm{Eyeglasses}$  & $\textrm{7.25}$    &$\textrm{Smiling}$ &3.36     \\
\hline
$\textrm{Heavy\_Makeup}$  & $\textrm{1.98}$   &$\textrm{Wearing\_Lipstick}$ &0.54     \\
\hline
$\textrm{High\_Cheekbones}$  & $\textrm{47.83}$   &$\textrm{Young}$  &0.49     \\
\hline
\end{tabular}
\end{table}

Using our methods, we can provide unique attribute description for multiple images of the same subject. We then check whether there is also inconsistency for the attributes labels (ground truth). Different from the previous procedures where the outputs are from deep features, this time we calculate their IM based on the attributes labels and the corresponding subjects. Following (7) , the IM is calculated for the annotated labels as shown in TABLE \uppercase\expandafter{\romannumeral6}.

% we make another experiment to question whether there is inconsistency issue on the attributes labels. Different from the procedures we conducted on PaSC and COX datasets, we first calculate their IM based on the attributes labels and corresponding identities, instead of deep feature classifier results. Following (7) , the IM is generated for the manual labels as shown in \textcolor[rgb]{1.00,0.00,0.00}{Table \uppercase\expandafter{\romannumeral5}}.

From TABLE \uppercase\expandafter{\romannumeral6}, we can see that the ground truth labels have the inconsistency issue. Excluding those dependent attributes, Arched\_Eyebrows, Pointy\_Nose, and Oval\_Face, etc. there is still a relatively high IM which indicates the inconsistency. Our proposed approach can handle this issue and correct the incorrectly annotated labels.
%Eventually, we are not only able to provide a certain consistent attribute, but another interesting contribution is that we can correct these manual incorrect labels using the method we proposed.

As we know, labeling data is expensive, but we do need these manual works to service better performance for deep learning. But how to find the label's correctness is difficult and expensive too. Taking the gender as an example, it would take massive human force to manually check the mistakes for gender annotation. Nonetheless, our method can be used to correct the errors, as shown in Fig.6. We can consider the highest confidence and quality or adopt the fusion idea as we described in Section \uppercase\expandafter{\romannumeral4}, and finally provide the consistent attribute labels.

%Ultimately, we conduct the solutions as we addressed, and compute the attributes accuracies as shown in \textcolor[rgb]{1.00,0.00,0.00}{figure}.

\begin{table*}[!htbp]
\centering
\caption{\label{table1}Label Inconsistence Measure on CelebA.}
\begin{tabular}{cc|cc|cc|cc|cc}
\hline
$\textbf{Attributes}$  & $\textbf{IM}$   &$\textbf{Attributes}$  & $\textbf{IM}$ &$\textbf{Attributes}$  & $\textbf{IM}$&$\textbf{Attributes}$  & $\textbf{IM}$&$\textbf{Attributes}$  & $\textbf{IM}$    \\
\hline
\hline
$\textrm{5\_o\_Clock\_Shadow}$  & $\textrm{11.31}$  &$\textrm{Black\_Hair}$    &27.99 &$\textrm{Goatee}$    &6.42   &$\textrm{No\_Beard}$    &11.70 &$\textrm{Straight\_Hair}$    &29.02  \\
\hline
$\textrm{Arched\_Eyebrows}$  & $\textrm{26.63}$  &$\textrm{Blond\_Hair}$    &13.53 &$\textrm{Gray\_Hair}$    &4.89   &$\textrm{Oval\_Face}$    &35.24 &$\textrm{Wavy\_Hair}$    &35.09  \\
\hline
$\textrm{Attractive}$  & $\textrm{31.76}$  &$\textrm{Blurry}$    &11.25 &$\textrm{Heavy\_Makeup}$    &25.18   &$\textrm{Pale\_Skin}$    &7.32 &$\textrm{Wearing\_Earrings}$    &28.22  \\
\hline
$\textrm{Bags\_Under\_Eyes}$  & $\textrm{28.65}$  &$\textrm{Brown\_Hair}$    &27.41 &$\textrm{High\_Cheekbones}$    &46.46   &$\textrm{Pointy\_Nose}$    &28.71 &$\textrm{Wearing\_Hat}$    &7.53  \\
\hline
$\textrm{Bald}$  & $\textrm{2.73}$  &$\textrm{Bushy\_Eyebrows}$    &16.42 &$\textrm{Male}$    &1.26   &$\textrm{Receding\_Hairline}$    &13.63 &$\textrm{Wearing\_Lipstick}$    &15.53  \\
\hline
$\textrm{Bangs}$  & $\textrm{18.93}$  &$\textrm{Chubby}$    &8.01 &$\textrm{Mouth\_Slightly\_Open}$    &55.50   &$\textrm{Rosy\_Cheeks}$    &11.19 &$\textrm{Wearing\_Necklace}$    &21.45  \\
\hline
$\textrm{Big\_Lips}$  & $\textrm{16.93}$  &$\textrm{Double\_Chin}$    &7.66 &$\textrm{Mustache}$    &4.77   &$\textrm{Sideburns}$    &6.66 &$\textrm{Wearing\_Necktie}$    &11.69  \\
\hline
$\textrm{Big\_Nose}$  & $\textrm{19.65}$  &$\textrm{Eyeglasses}$    &8.79 &$\textrm{Narrow\_Eyes}$    &23.53   &$\textrm{Smiling}$    &52.77 &$\textrm{Young}$    &6.71  \\
\hline
\end{tabular}
\end{table*}

\begin{figure*}[htb]
\begin{center}
\includegraphics[width=1.0\textwidth]{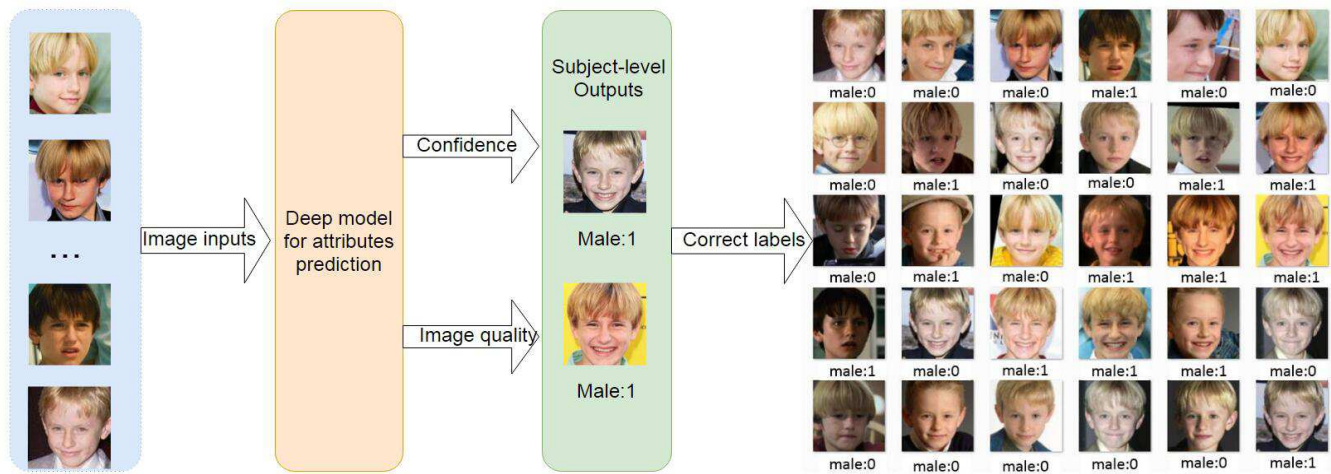}

\caption{\label{Fig.6} The right side shows the attribute labels of one identity on CelebA \cite{liu2015deep}. Even though they are from the same identity, the attribute label (Male) has encountered inconsistency. Using our methods, two representations are selected based on confidence and image quality, and we can output the subject-level attribute estimation to correct the incorrectly annotated labels.}
\end{center}
\end{figure*}

\section{CONCLUSION}

In this work, we proposed a novel problem to study and developed methods for facial attributes from multiple images of the same subject. We illustrated the face attributes inconsistence issue when dealing with multiple images or video frames. After that, we developed two approaches to address the problem using probabilistic confidence and image quality assessment. Given these approaches, the unique facial attribute can be computed. Moreover, our methods can be applied to correct the incorrectly annotated labels in a large database.
%Certainly, all these attributes prediction results are based on our ensemble deep model for extracting feature and efficient classifiers, which have achieved a comparable performance compared to the state of the art \cite{rudd2016moon}.

%This is a repeat.
%Position figures and tables at the tops and bottoms of columns.
%Avoid placing them in the middle of columns. Large figures and tables
%may span across both columns. Figure captions should be below the figures;
% table captions should be above the tables. Avoid placing figures and tables
%  before their first mention in the text. Use the abbreviation ``Fig. 1'',
%  even at the beginning of a sentence.
%Figure axis labels are often a source of confusion.
%Try to use words rather then symbols. As an example write the quantity ``Inductance",
% or ``Inductance L'', not just.
% Put units in parentheses. Do not label axes only with units.
% In the example, write ``Inductance (mH)'', or ``Inductance L (mH)'', not just ``mH''.
% Do not label axes with the ratio of quantities and units.
% For example, write ``Temperature (K)'', not ``Temperature/K''.

%%%%%%%%%%%%%%%%%%%%%%%%%%%%%%%%%%%%%%%%%%%%%%%%%%%%%%%%%%%%%%%%%%%%%%%%%%%%%%%%
\section{ACKNOWLEDGMENTS}
This work is partly supported by a NSF-CITeR grant and a WV HEPC grant.

%%%%%%%%%%%%%%%%%%%%%%%%%%%%%%%%%%%%%%%%%%%%%%%%%%%%%%%%%%%%%%%%%%%%%%%%%%%%%%%%

%\begin{thebibliography}{99}
%
%\bibitem{c1}
%J.G.F. Francis, The QR Transformation I, {\it Comput. J.}, vol. 4, 1961, pp 265-271.
%
%\bibitem{c2}
%H. Kwakernaak and R. Sivan, {\it Modern Signals and Systems}, Prentice Hall, Englewood Cliffs, NJ; 1991.
%
%\bibitem{c3}
%D. Boley and R. Maier, "A Parallel QR Algorithm for the Non-Symmetric Eigenvalue Algorithm", {\it in Third SIAM Conference on Applied Linear Algebra}, Madison, WI, 1988, pp. A20.
%
%\end{thebibliography}

\bibliographystyle{ieeetr}
\bibliography{myBibTex}

\end{document}